\documentclass[conference]{IEEEtran}
\IEEEoverridecommandlockouts
\ifCLASSINFOpdf
\usepackage[pdftex]{graphicx}
\usepackage{footnote}
\usepackage{booktabs}
\else
\fi
\usepackage{amsmath}
\begin{document}
%
\title{Empirical Studies on Symbolic Aggregation Approximation Under Statistical Perspectives for Knowledge Discovery in Time Series }

\author{\IEEEauthorblockN{Wei Song 
}
\IEEEauthorblockA{School of Information Engineering\\
Zhengzhou University\\
iewsong@zzu.edu.cn
}
\and
\IEEEauthorblockN{Zhiguang Wang 
}
\IEEEauthorblockA{Dept.of CS and EE\\
University of Maryland \\
Baltimore County \\
stephen.wang@umbc.edu
}
\and
\IEEEauthorblockN{Yangdong Ye, Ming Fan}
\IEEEauthorblockA{School of Information Engineering\\
Zhengzhou University\\
ieydye,iemfan@zzu.edu.cn
}
}


%


\maketitle

\begin{abstract}
Symbolic Aggregation approXimation (SAX) has been the de facto standard representation methods for knowledge discovery in time series on a number of tasks and applications. So far, very little work has been done in empirically investigating the intrinsic properties and statistical mechanics in SAX words. In this paper, we applied several statistical measurements and proposed a new statistical measurement, i.e. information embedding cost (IEC) to analyze the statistical behaviors of the symbolic dynamics. Our experiments on the benchmark datasets and the clinical signals demonstrate that SAX can always reduce the complexity while preserving the core information embedded in the original time series with significant embedding efficiency. Our proposed IEC score provide a priori to determine if SAX is adequate for specific dataset, which can be generalized to evaluate other symbolic representations. Our work provides an analytical framework with several statistical tools to analyze, evaluate and further improve the symbolic dynamics for knowledge discovery in time series.
\end{abstract}


\textit{\textbf{Key Words---} SAX; Knowledge Discovery in Time Series; Infomation Embedding Cost; Permutation Entropy; Symbolic Complexity}

%
\IEEEpeerreviewmaketitle

\let\thefootnote\relax\footnotetext{This work is supported by the National Natural Science Foundation of China (Grant No.61170223), The Education Department of Henan Province (Grant No.13A520453), Department of Science \& Technology of Henan Province (Grant No.142300410229).}

\section{Introduction}
Time series is a sequence of data obtained from sequential measurements over time. Nowadays, time series data is ubiquitous among finance, health care, multimedia, agriculture and manufacturing, etc. Arising needs on various time series data mining tasks inspire a number of representation methods to reduce the volume, smooth the noise or extract the implicit structure for further data mining tasks. Some numerical approaches inspired by signal processing techniques are firstly applied. The Discrete Fourier Transformation (DFT) framework is applied in the seminal work of Agrawal \cite{agrawal1993efficient} by projecting the time series into the sine and cosine bases. Discrete Wavelet Transformation (DWT), which uses the scaled variety of mother wavelet function to give multiresonlutional decomposition basis to measure both high frequency and low frequency over large intervals, are intensively applied in the literature over last decades \cite{grinsted2004application,lau1995climate,ivanov1996scaling}. Singular Value Decomposition (SVD) and its extensions have been proposed to find multiscale patterns in the streaming data \cite{korn1997efficiently,ravi1998dimensionality}. Esling and Agon had a good review paper about other numerical representation methods \cite{esling2012time}.

Instead of generating a sequence of numerical output, symbolic representations provide another perspective. Aligned Cluster Analysis (ACA) is introduced as an unsupervised method to cluster the temporal patterns of human motion data \cite{zhou2008aligned}. It is an extension of kernel k-means clustering but requires quite computational capacity. Persist is an unsupervised discretization methods to maximize the persistence measurement of each symbol \cite{morchen2006finding}.  Piecewise Aggregation Approximation (PAA) methods is proposed by Keogh \cite{keogh2001dimensionality} to reduce the dimensionality of time series, which is then upgraded to Symbolic Aggregation Approximation (SAX) \cite{lin2003symbolic}. In SAX, each aggregation value after PAA process is mapped into the equiprobable intervals based on standard normal distribution to produce a sequence of symbolic representations. Sant'Anna et al. compared the comprehensive performance of above three symbolic approaches \cite{sant2011symbolization}.

Among the above representation approaches, SAX method has become one of the de facto standard to discretize time series and is at the core of many effective classification algorithms. Koegh et al. introduce the new problem of finding time series discords and apply SAX derivations to find the subsequences of a longer time series that are maximally different to all the rest of the time series subsequences \cite{keogh2005hot}. A new SAX-based algorithm is proposed to discover time series motifs with invariance to uniform scaling. The authors show that their approach produces objectively superior results in several important domains \cite{yankov2007detecting}.  Vector Space Model is combined together with SAX as a novel method to discover characteristic patterns in a time series \cite{senin2013sax}. SAX with bag-of-words are proved to be powerful in several time series classification tasks \cite{oates2012predicting,oates2012exploiting,wang2014time}.

As the golden standard of supervised parametric approach in recent ten more years, what are the intrinsic statistical behavior of SAX? How can we evaluate and justify its performance through empirical studies? What can we learn from SAX to further improve the time series representations? The following properties/questions primarily motivate our work:

\begin{itemize}
\item Data complexity: Time series data are inherently noisy and chaotic. Symbolization can improve the analysis through the assumption that it compresses the complexity and generates more concise representations.
\item Information loss: The PAA process of SAX reduces the volume but always incurs the information loss. The ideal results are achieved when the information loss is superfluous, e.g. noise is abandoned in the analysis of the raw time series.
\item Efficiency of information embedding: While SAX is supposed to smooth the noise, the following question is natural to ask about, how much useful information in original signals will be integrated into the symbolic output of SAX?
\item Seasonality and correlation: correlation and seasonality is important property in time series analysis, e.g. to determine the order of ARIMA model. For pattern discovery task, explicitly extract the implicit seasonality in time series helps us to reveal the intrinsic correlations. However, redundant correlation will affect the representation efficiency. 
\end{itemize}

We empirically studied above four aspects under a statistical perspectives. Briefly, permutation entropy \cite{bandt2002permutation} helps us to reveal the implicit complexity under the seemingly randomness or chaos. KL divergence \cite{kullback1951information} measures the distribution distance between symbolic output and the raw data. The information reduction/loss is quantified by standard reconstruction information loss and a new measurement, Information Embedding Cost (IEC) which is built upon the standard reconstruction information loss and KL divergence are applied to analyze the information embedding efficiency. We consider Autocorrelation function (ACF) and partial autocorrelation function (PACF) \cite{wei1994time} as well to deeply understand the seasonality and internal correlation embedded in the raw data and symbolic outputs. We analyzed above properties between SAX, PAA and raw data on seven benchmark datasets \cite{keogh2006ucr} and the real world clinical signals. Our proposed analysis framework shows concretely reasonable conclusions.

\section{background}
In this section, we briefly introduce the SAX approach. SAX can be manually divided into two phases to facilitate our discussion in this paper. First, we reduce the dimensionality of the input time series by  Piecewise Aggregation Approximation (PAA). Each value is the mean of its corresponding sliding windows. Then PAA values are mapped to the equiprobable intervals according to standard normal distribution as alphabet characters. Overall time series trend will be extracted as a symbolic sequence as SAX words.

\begin{figure}[t]
\centering
\includegraphics[width=3.3in]{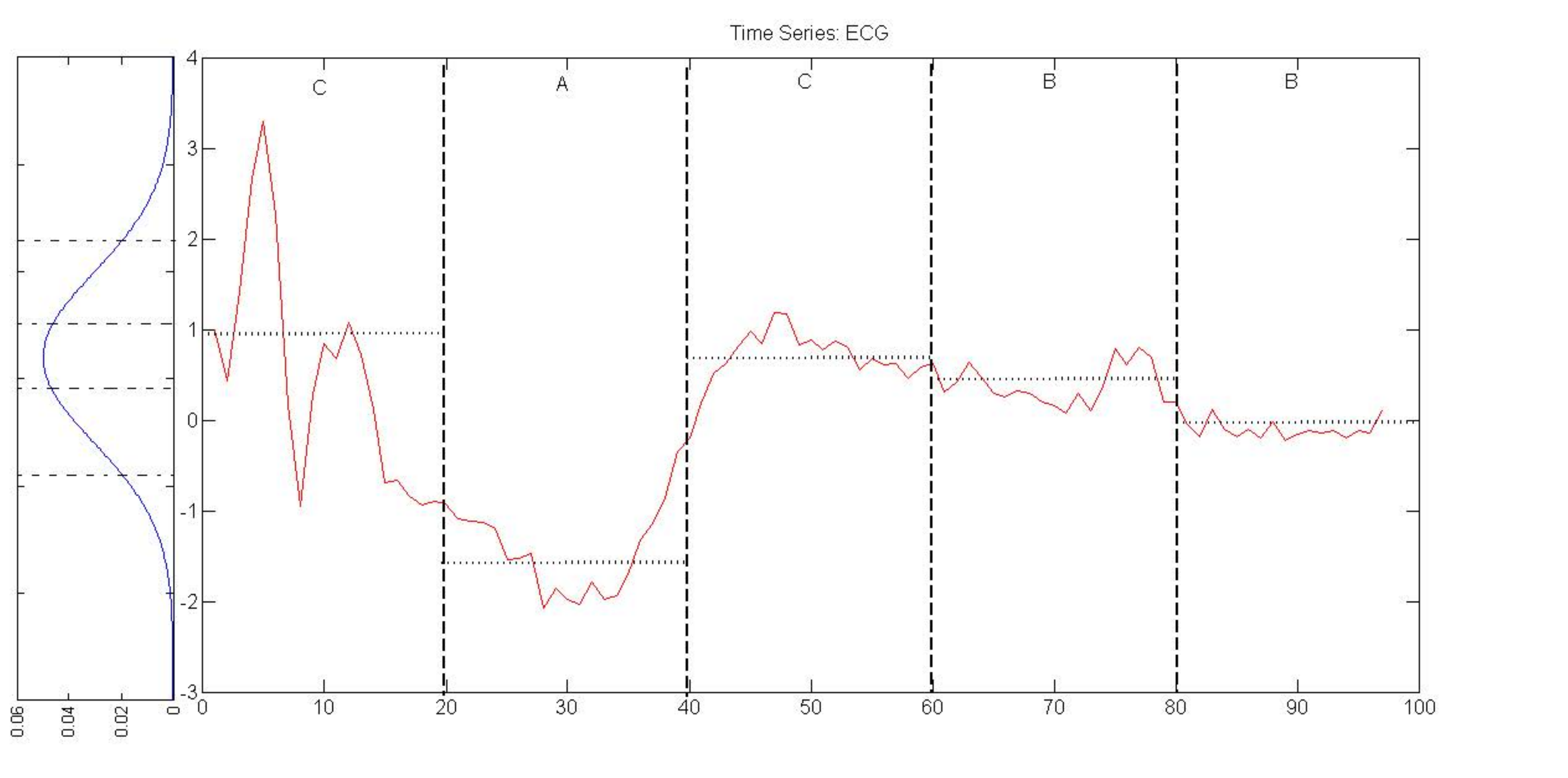}
\caption{The piecewise aggregate approximation for the "ECG" data and the corresponding SAX words.}
\label{fig:SAX-demo}
\end{figure}

The algorithm has three parameters: window length $n$, word number $w$ and alphabet size $a$. Different parameters lead to different representations of the time series. Given a normalized time series of length $n$, in the first step we need to reduce the dimensionality by dividing it into $[n/w]$ non-overlapping subwindows. Mean values of each sub-window are computed to reduce volume and smooth (PAA procedure). Then, PAA values are mapped to a probability density function $\mathcal{N}(0,1)$ to divide the PAA value into equiprobable segments with the same probabilities. Words starting from $A$ to $Z$ are assigned to each PAA values corresponding to the segments they fall in. Fig. \ref{fig:SAX-demo} shows the PAA and SAX word of ECG signals from the UCR dataset. A time series of length 96 is partitioned into 5 segments. The means of each segment are allocated to the equiprobable interval. After discretization by PAA and symbolization by SAX, we convert the time series into symbolic sequence $CACBB$.

\section{methods}
\subsection{Permutation Entropy (PE)}
Permutation entropy (PE) is a measurement in symbolic dynamics to quantify the complexity of time series. Given a sequence of time series $\{x_t\}_{t=1...T}$, PE considers all the permutation sets $\pi$ of order length $n$. All the possible orders of $n$ different numbers are denoted as $S_n$. For each $\pi$ in $S_n$, we determine the relative frequency of possible permutation patterns as:
\begin{equation}
p(\pi)=\frac{\#\{0 \leq t \leq T-n, \{x_{t+1},x_{t+2},...,x_{t+n}\} \in \pi \}}{T-n+1}
\end{equation}

When the order $n \geq  2$, PE value is defined as
\begin{equation}
H(n)=-\sum P(\pi)\log P(\pi)
\end{equation}

PE value is actually the sum runs over all possible $n!$ permutations in permutation pool $S_n$. It is clear that $0 \leq H(n) \leq \log n!$. We use the normalized PE $\frac{H(n)}{\log n!}$ to scale the PE value into $[0,1]$ to facilitate our comparison.

PE has two parameters. Permutation order $n$ determines the length of the sequence, \textit{i.e.} the size of the permutation pools. Time delay $t$ controls the time embedding properties between successive points in the symbol sequences. The optimal $n$ and $t$ are highly dependable on the system when using PE to determine the optimal embedding parameters \cite{riedl2013practical}. We discuss the comparable complexity of raw data, PAA values and SAX words instead of finding a group of optimal parameters to precisely quantify the complexity. To simplify, we set $t=1 -100$ to observe the tendency of PE values by changing the time delay. For the parameter $n$, we follow the suggestions to constrain $N>5n!$ \cite{amigo2008combinatorial}, thus $n \leq 7$.  Values with the same rank number in the regarded sequence generate a new permutation pattern, \textit{e.g.}, 3, 4, 4, 3, 1 leads to the permutation pattern 1, 2, 2, 1, 0 \cite{bian2012modified}.

\subsection{Information Embedding Cost (IEC)}
\subsubsection{information loss}
Discretization always incur information loss. Piecewise averaging by PAA and SAX will reduce the dimensionality, but also ignore specific low frequency details in the raw time series. Information loss is estimated by Mean Square Error (MSE) between original signal and reconstructed symbolic sequences, which is slightly different from \cite{sant2011symbolization}:

\begin{equation}
\text{information loss }(\tilde{T},T)=\frac{\sum (\tilde{t_i}-t_j)^2}{n-1}, \tilde{t_i} \in \tilde{T}, t_j \in T
\end{equation}

$T$ is the original signal and $\tilde{T}$ is the reconstructed signal. Reconstructed signal was generated by substituting the original samples with its corresponding PAA values. For SAX, piecewise averaging from normalized signals are mapped to its corresponding equiprobable intervals for symbolization. Then the original samples were substituted by its corresponding SAX words. To facilitate the comparison and analysis, raw time series, PAA values and SAX words (A-Z correspond to digits 0-25) are scaled to $[0,1]$ (Fig. 2).

\begin{figure}[t]
\centering
\includegraphics[width=2.5in]{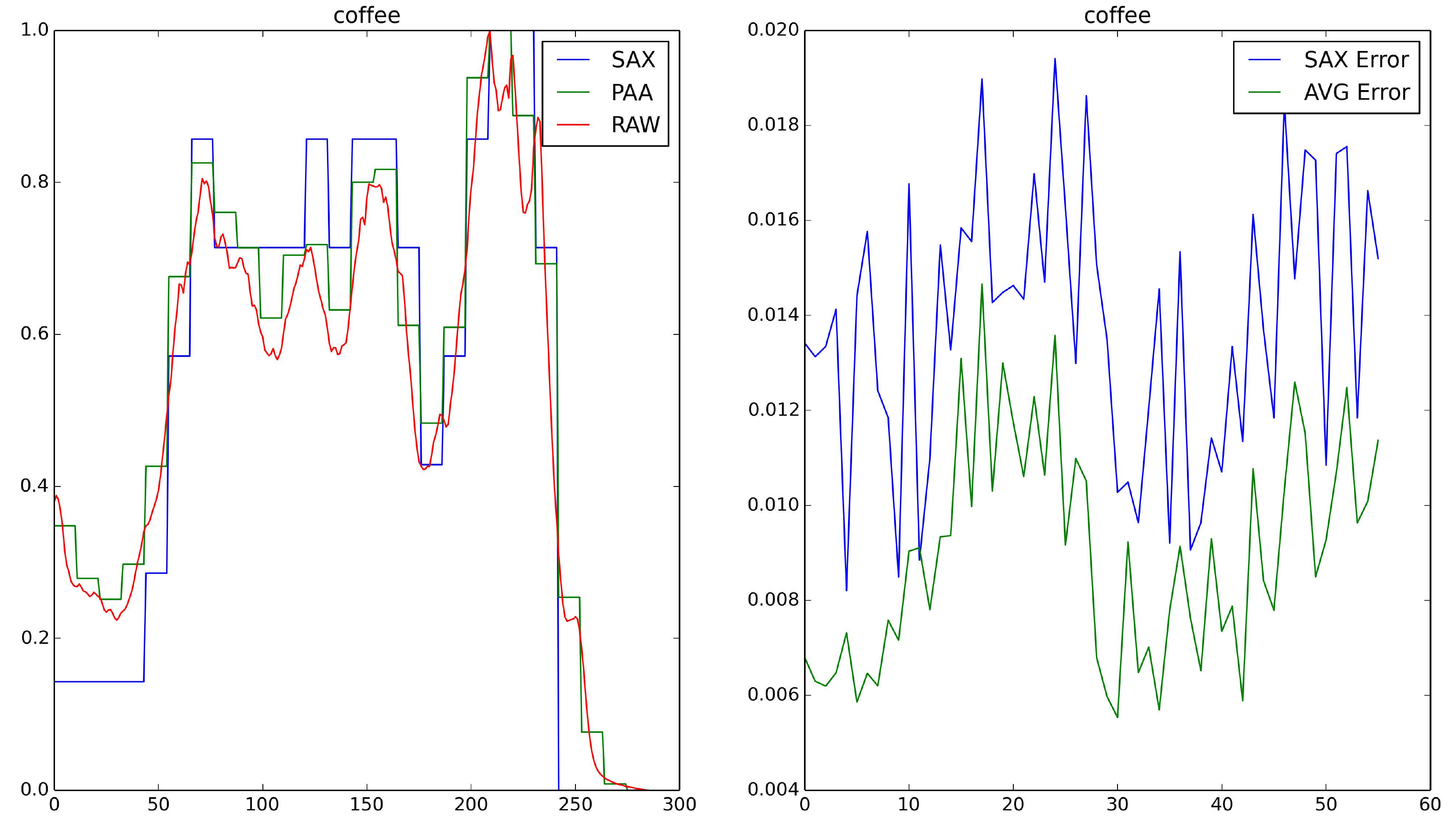}
\caption{Reconstructed signals of SAX word and PAA representation (left) and MSE over all samples on UCR Coffee dataset.}
\label{fig:info_loss_sample}
\end{figure}

\subsubsection{KL divergence}
A measurement to compare the distance between two different probability distributions. For distributions $P$ and $Q$ of $k$ points, the non-symmetric KL divergence is defined as:

\begin{equation}
KL(P||Q)=\sum_{i=1}^{k}p_i\log \frac{p_i}{q_i}, p_i \in P, q_i \in Q
\end{equation}
In information theory, the KL divergence measures the expected number of extra bits required to code the samples from $P$ when using a code based on $Q$ rather than using the original code based on $P$ itself. In the our experiments, $P$ is the distribution of quantile bins in the original signals, the number of bins actually equals to the alphabet size $a$ in SAX/PAA settings.

Information loss measures the amount of information discarded when coding the original signal to the symbolic output. KL divergence indicates the distance between the distribution from encoding outputs to the original signals. We define IEC score as the ADD-1 ratio of standardized KL divergence and standardized information loss. IEC measures how much useful information is discarded when approximating the original signals:

\begin{equation}
IEC_T(P,Q)=\frac{KL_T(P||Q)_{std}}{1+InfoLoss(\tilde{T},T)_{std}}
\end{equation}

Given time series $T$ in distribution $Q$ and the proposed encoding approach $\phi$ with distribution $P$, its IEC score demonstrates the number of extra bits required when encoding output $\tilde{T}$ when one unit information loss incurs. Note that we do not equally emphasis on KL divergence and information loss in the IEC framework. The IEC score linearly decreases as KL divergence falling. However, when information loss increases, IEC score not only corresponds to the inverse square term $\frac{-1}{(1+InfoLoss_{(\tilde{T},T)})^2}$,  but also depends on the value of KL divergence. This means that IEC score emphasis more on the KL divergence, or the information embedding accuracy rather than the compression rate of symbolic representations. Intuitively, this is reasonable because we need precise representations rather than the symbols that are too concise to preserve enough information from the original signals.

$IEC=0$ when the encoding approach $\phi$ generates the representations with the same probability distribution as the original signals. $IEC=1$ if and only if the deviation measured by KL divergence reaches to the maximum, \textit{i.e.} no bit of original signal is compressed by the proposed symbolic representation. $IEC=0.5$ when KL divergence and information loss are both 1. It is the baseline when applying IEC score to evaluate the information embedding efficiency of specific symbolic representation. $IEC=0.5$ means the symbolic representation will lose one unit information while reduce one unit complexity. Relatively large information loss and small KL divergence demonstrate that the encoding method preserves much information while reducing the information complexity. Thus, the information loss caused by encoding method with small IEC score is more likely to produce more concise and accurate representation.

\subsection{(Partial) Autocorrealtion Coefficient Function}
For a stationary process $Z={Z_1,Z_2,…,Z_t}$, the covariance between $Z_t$ and $Z_{t+k}$ is defined as
\begin{equation}
\gamma_k=cov(Z_t,Z_{t+k})=E(Z_t-\mu)(Z_{t+k}-\mu)
\end{equation}

The correlation between $Z_t$ and $Z_{t+k}$ as:
\begin{equation}
\rho_k=\frac{\gamma_k}{\sqrt{var(Z_t)}\sqrt{var(Z_{t+k})}}
\end{equation}

As a function of $k$, $\rho_k$ is the Autocorrelation Coefficient Function (ACF). ACF represents the covariance and correlation between $Z_t$ and $Z_{t+k}$ with the time lag $k$.

If we remove the mutual linear dependency on the intervening variables $Z_{t+1},Z_{t+2},...,Z_{t+k-1}$, the conditional correlation is described by:
\begin{equation}
Corr(Z_t,Z_{t+k}|Z_{t+1},Z_{t+2},...,Z_{t+k-1})
\end{equation}
Above equation is Partial Autocorrelation Coefficient Function (PACF). 

ACF and PACF are widely applied in time series analysis to identify the order of the ARIMA model. We focus on the essential property of ACF and PACF, \textit{i.e.} the seasonality of time series. Symbolic representation reduces the complexity of the original signal while running the risk of information loss and information distortion. Regular patterns such as seasonality extracted from noisy and chaotic data is important in discovering more reasonable patterns embedded in the original data, but over-complicated correlation leads to large information redundancy. The ideal representation removes much of the superfluous correlation while retaining the main invariant dependencies.

\section{experiments and analysis}
We select 7 benchmark datasets from the UCR Time Series Classification and Clustering Repository \cite{keogh2006ucr}. We also consider the real clinical signals to validate our results. Our clinical data were collected in University of XXX School of Medicine. All patient data are anonymous in order to protect privacy. 556 patients’ ECG and PPG data were collected in 68 to 128 minutes long with 240 Hz sampling rate. SAX requires three parameters, window length $n$, word number $w$ and alphabet size $a$. To empirically explore the intrinsic property of the appropriate SAX representations, we firstly use SAX-BoP representations with a 1NN classifier to classify the benchmark datasets, as described in \cite{oates2012exploiting}. We extracted the average value of the alphabet size in the top 10 representations for each dataset. For simplicity, Window length $n$ is the sequence length, word number $w$ and alphabetical size $a$ are the rounding values of the expectation among the top 10 representations. All the parameters are shown in Table \ref{tab:SAX-param}.

\begin{table}[t]
  \centering
  \caption{Optimal SAX parameters on seven benchmark datasets.}
    \begin{tabular}{rrrr}
    \toprule
    \textbf{Dataset} & \textbf{Length n} & \textbf{Rounding w} & \textbf{Rounding a} \\
    \midrule
    ECG & 96 & 12 & 7 \\
    Lighting2 & 637 & 18 & 7 \\
    Coffee & 286 & 48 & 7 \\
    Adiac & 176 & 25 & 9 \\
    Lighting 7 & 319 & 11 & 9 \\
    Beef & 470 & 11 & 5 \\
    Oliveoil & 570 & 26 & 7 \\
    \bottomrule
    \end{tabular}%
  \label{tab:SAX-param}%
\end{table}%

\subsection{Complexity}

\begin{figure}[t]
\centering
\includegraphics[width=3.5in]{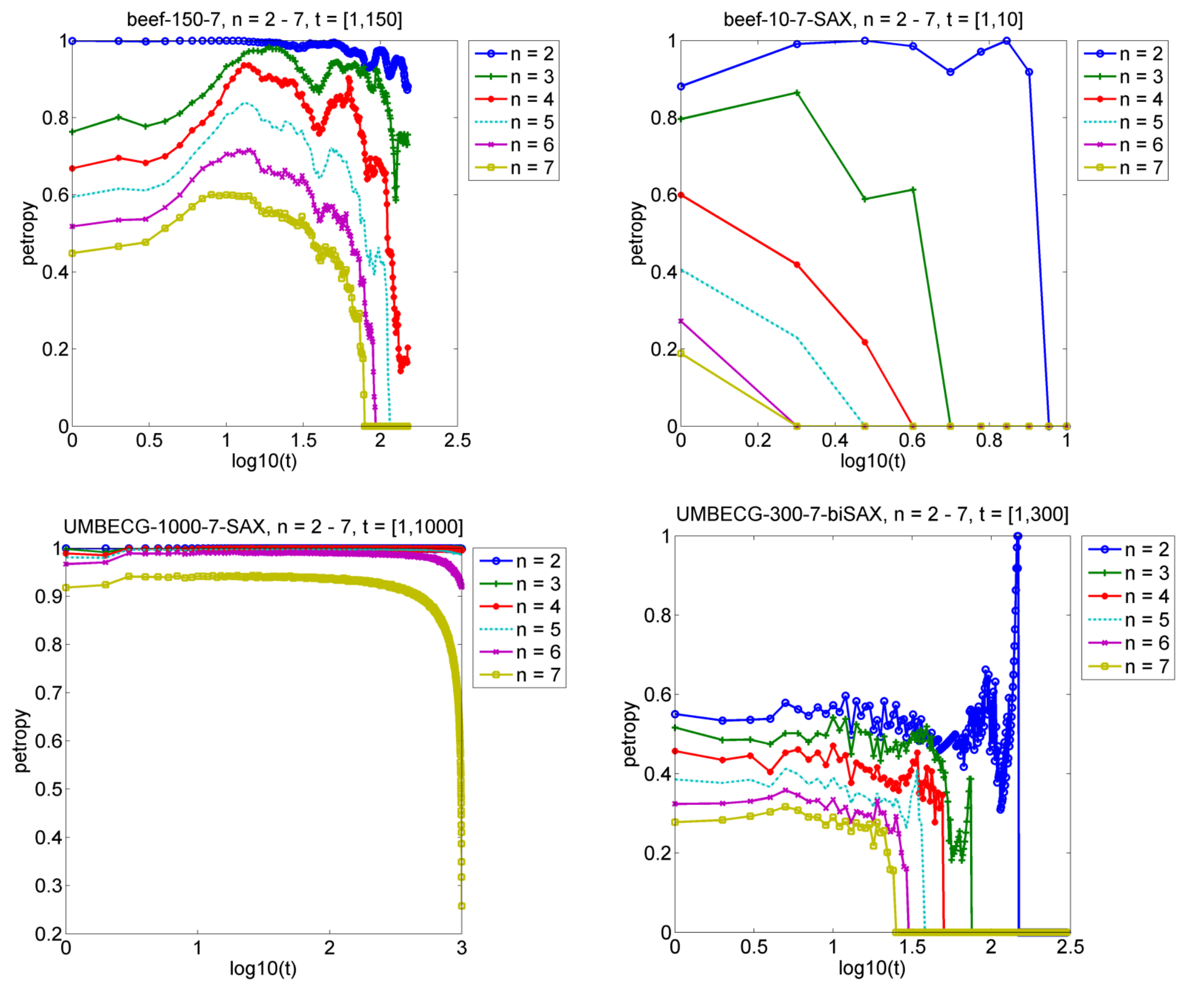}
\caption{Permutation entropy on (a) raw UCR time series, permutation order $n = [2,7]$, $t$ is up to $\frac{1}{3}$ of the length (up left),(b) SAX words from UCR time series, permutation order $n = [2,7]$, $t =  [1,10]$ (up right), (c) clinical signals (bottom left), (d) SAX words on clinical signals (bottom right).}
\label{fig:PE-graph}
\end{figure}

To study the representational complexity of SAX, we consider the normalized permutation entropy on the benchmark datasets. For the clarification and the sake of space, we show the graph upon one datasets. However, our analysis are more general on all the datasets. In Table \ref{fig:PE-graph} (1), 
it is clear that a linear rise of the PE value incurs for the increasing $t$. It demonstrates the complexity of the unfolding trajectory is also increased. The correlation between the values in the embedding vector $t \times (n-1)$ is getting lost. This observation is named 'redundancy effect' by DeMicco \textit{et. al} \cite{de2012sampling}. High correlation leads to less need of visiting the phase of the data during the reconstruction of the trajectory. The increasing PE value shows the data is correlated within the embedding vector of $t \times (n-1)$. The embedding vector is too small for the original data to summarize the full information among the potentially successive values in the time series that fall into the embedding vectors. The correlation, or the embedded information between the original time series is loose with much more noise and redundant irrelevance. This assumption is clearly supported by the observation, that along within the embedding vector $t \times (n-1)$, the PE value is increasing linearly without convergence in Figure \ref{fig:PE-graph} (a). Relatively high internal correlation leads to information redundancy in high dimensionality of time series and enhances the curse of dimensionality \cite{bellman1961adaptive}.

After symbolization by SAX, the redundancy effect is significantly alleviated (Figure \ref{fig:PE-graph} (b)). However, this may be caused by two possible reasons. Note that the symbolic sequence is shorter in length, one reason is that the number of permutation patterns is limited as $t$ increases. Another potential reason is that SAX word not only decreases the dimensionality, it also makes the correlation more 'compact'. Instead of the data with very loose correlations, the SAX words does not experience in heavy redundancy effect when it reaches to its convergence or starts to diffuse.

Moreover, the magnitude of the PE values demonstrates the absolute complexity. Compare (a) and (b) in Figure \ref{fig:PE-graph} (and on all datasets), the magnitude of the PE values overlap between raw and SAX However, in the benchmark dataset, the time series is not very noisy and the information is compact. We compared the PE values between the raw clinical data and its corresponding SAX words (Figure \ref{fig:PE-graph} (c) and (d)), SAX words clearly has lower PE magnitude than the raw data . Thus, SAX practically reduces the redundancy effect and the absolute complexity.

\subsection{Information Embedding Efficiency}
\begin{figure}[t]
\centering
\includegraphics[width=3.5in]{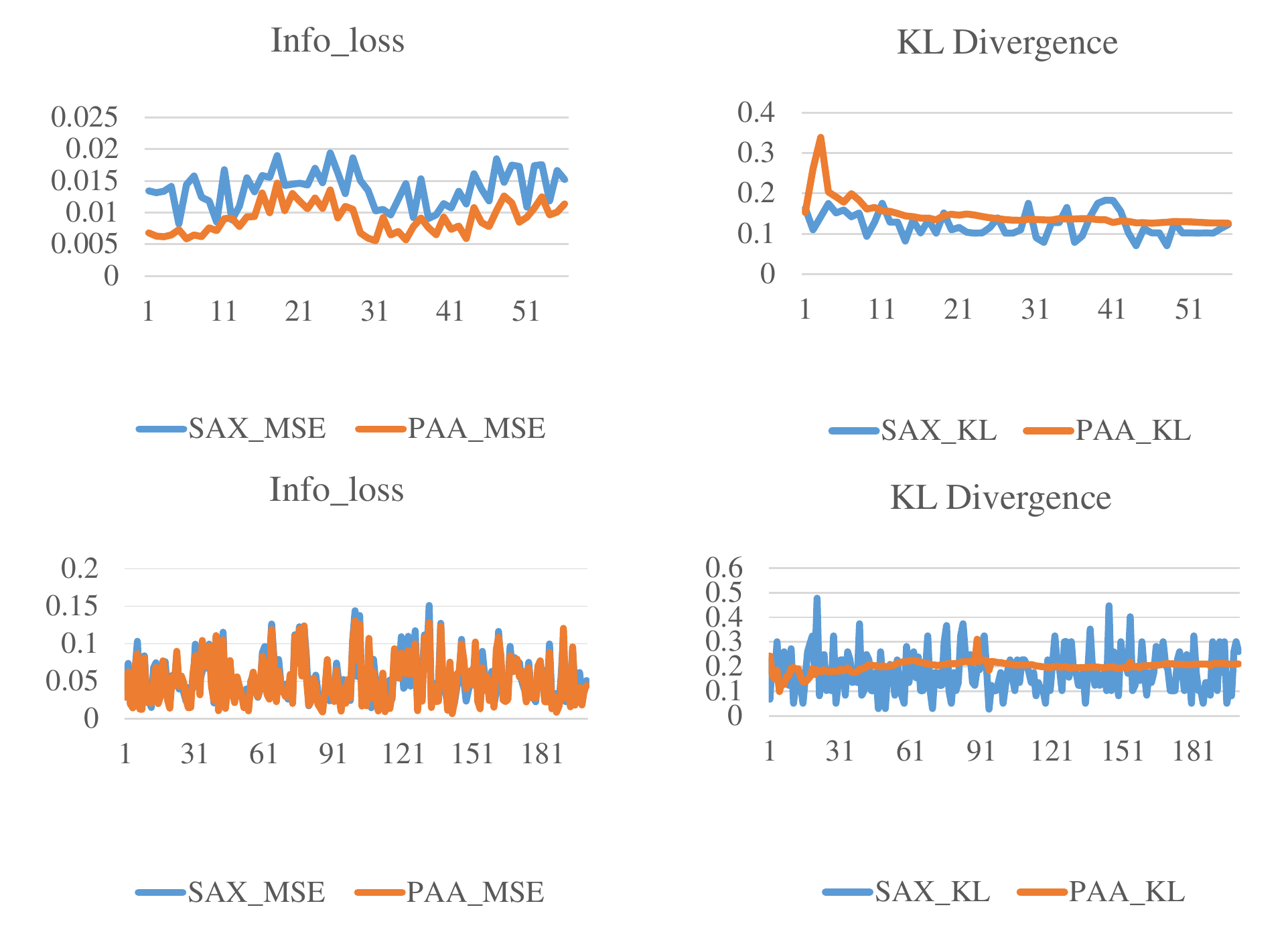}
\caption{Information loss and KL divergence on (a) 'Coffee' dataset (up), (b) 'ECG' dataset (bottom). X-axis denotes the sample index, Y-axis is the corresponding measurement.}
\label{fig:KL-infor-UCR}
\end{figure}

As the size of SAX representation sometimes exerts subtle influence on the analysis based on PE, explore other ways to study the information embedding efficiency is necessary. 

SAX steps further than PAA by mapping the PAA values to the corresponding words according to the standard Gaussian distribution. If SAX can encode more information in the SAX words while incurs less information loss, we suppose to study if this advantage benefits from PAA process or Gaussian mapping process. However, neither KL divergence nor information loss are capable to analyze the non-linear dependencies among different dataset, but run into a case-by-case manner. On some datasets, it is easy to discriminate the benefit from the SAX words over the PAA output (Figure \ref{fig:KL-infor-UCR}a). However, the relative relationship might be ambiguous neither (Figure \ref{fig:KL-infor-UCR}b). If a representation has small KL divergence with superior information loss, it will maintain more substantial information while discarding much of the noise. Thus, to compare the information embedding efficiency among different symbolic approaches, we propose the newly IEC score to measure how much information is discarded when compress the original signals. We take the average of IEC scores among all samples in each dataset and compare the classification performance of SAX \cite{oates2012exploiting} and original time series \cite{keogh2006ucr} using a 1NN classifier (Table \ref{tab:SAX-IEC}). We expect that the representations with lower IEC score achieve lower classification error rates.

\begin{table}[t]
  \centering
  \caption{The IEC scores on the SAX/PAA representations and the classification error rates on SAX words and the original data.}
    \begin{tabular}{crrrr}
    \toprule
    \textbf{Dataset} & \multicolumn{1}{c}{\textbf{SAX IEC}} & \multicolumn{1}{c}{\textbf{PAA IEC}} & \multicolumn{1}{c}{\textbf{Error Rate }} & \multicolumn{1}{c}{\textbf{Error Rate }} \\
    \textbf{} & \multicolumn{1}{c}{\textbf{}} & \multicolumn{1}{c}{\textbf{}} & \multicolumn{1}{c}{\textbf{on SAX}} & \multicolumn{1}{c}{\textbf{on Raw Data}} \\
    \midrule
    lighting2 & 0.444 & 0.179 & 0.229 & 0.197 \\
    ECG & 0.259 & 0.305 & 0.22 & 0.11 \\
    coffee & 0.126 & 0.242 & 0.107 & 0.25 \\
    adiac & 0.096 & 0.089 & 0.383 & 0.407 \\
    lighting7 & 0.496 & 0.065 & 0.397 & 0.37 \\
    beef & 0.416 & 0.407 & 0.466 & 0.4 \\
    oliveoil & 0.119 & 0.883 & 0.166 & 0.233 \\
    clinical ECG & 0.192 & 0.611 & 0.173 & 0.545 \\
    \bottomrule
    \end{tabular}%
  \label{tab:SAX-IEC}%
\end{table}%

\begin{figure}[b]
\centering
\includegraphics[width=3.5in]{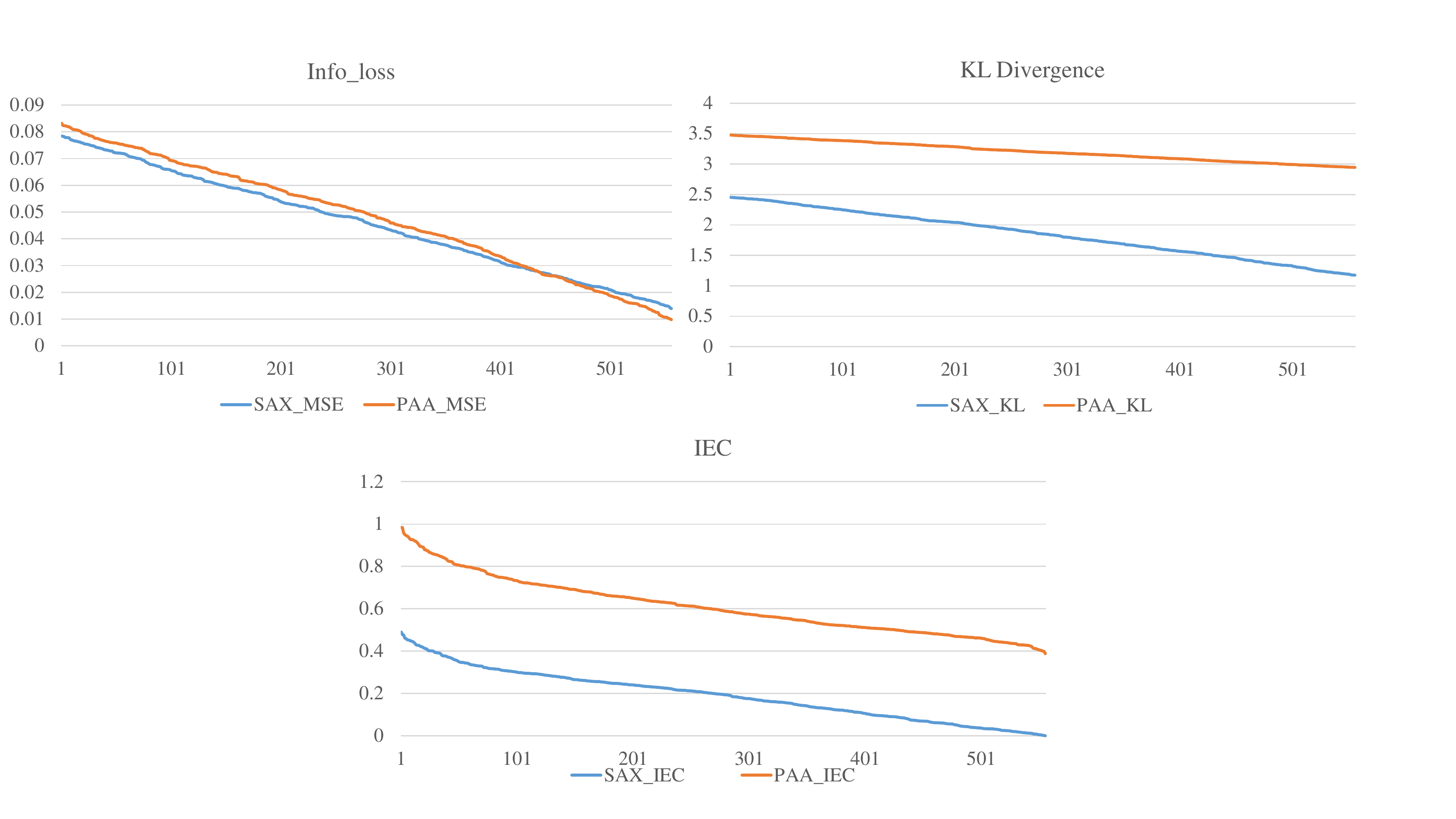}
\caption{The KL divergence, reconstruction information loss and IEC scores of the SAX words and PAA output on the clinical ECG data.}
\label{fig:KL-infor-clinical}
\end{figure}

Table \ref{tab:SAX-IEC} demonstrates that SAX words always has lower IEC scores than the PAA values. The Gaussian mapping procedure actually improves the representational capability by getting rid of much noise while maintaining the output symbols “nearer” to the original signals. Another observation is that, when SAX words has higher IEC score on some datasets (\textit{e.g.} 'lighting2' and 'lighting7'), the classification performance on SAX words are always worse than classifying the raw data. Actually, we suppose that the classification performance on PAA values will overtake SAX on the 'lighting2' and 'lighting7' datasets, because PAA values obviously has lower IEC scores than SAX words on these two datasets. On the clinical ECG dataset, the KL divergence, reconstruction information loss and IEC score of SAX words and PAA outputs clearly justify our assumptions. We rank above three measurements in ascending order (Figure \ref{fig:KL-infor-clinical}). Although its hard to judge the performance through the graph of KL divergence and information loss, obviously the SAX words has lower average IEC score over all samples (0.1919) than PAA output(0.6105). The low IEC score on SAX words also explains its superior performance  on classification comparing with the raw data (Table \ref{tab:SAX-IEC}).

To deeply understand the non-linear correlation encoded between the IEC scores and the representational capability for time series data mining (\textit{i.e. classification performance}), we applied the high-order regression model. Our assumption is that, the IEC score helps us to find out if one symbolic representation outperform raw data on specific dataset. We suppose that, the improvement on classification error rate of the SAX words over the original signal, defined as $\frac{ERR_{SAX}}{ERR_{raw}}$, are correlated to their corresponding IEC scores. By the quadratic regression, we found the error ratio and the IEC scores are highly correlated with the a R-square 0.9265 (Table. 3). Thus, we suggest that when the IEC score on the symbolic representation is lower than 0.2 with some specific tolerance, it is more likely to overtake the original signals in the data mining tasks such as classification or retrieval.

\begin{table}[t]
  \centering
  \caption{Regression statistics of the classification error ratio and the IEC scores on seven benchmark datasets and one real-world clinical dataset. Independent variable $x$ is the IEC score, dependent variable $y$ is the error ratio.}
    \begin{tabular}{rr}
    \toprule
    \textit{Regression Statistics} & \textit{} \\
    \midrule
    Multiple $R$ & 0.962576 \\
    $R$ Square & 0.926553 \\
    Intercept & 0 \\
    $x$ & 9.454057 \\
    $x^2$ & -14.9802 \\
    \bottomrule
    \end{tabular}%
  \label{tab:IEC-fit}%
\end{table}%

\subsection{Internal Correlation/Seasonality}

\begin{figure}[b]
\centering
\includegraphics[width=3.3in]{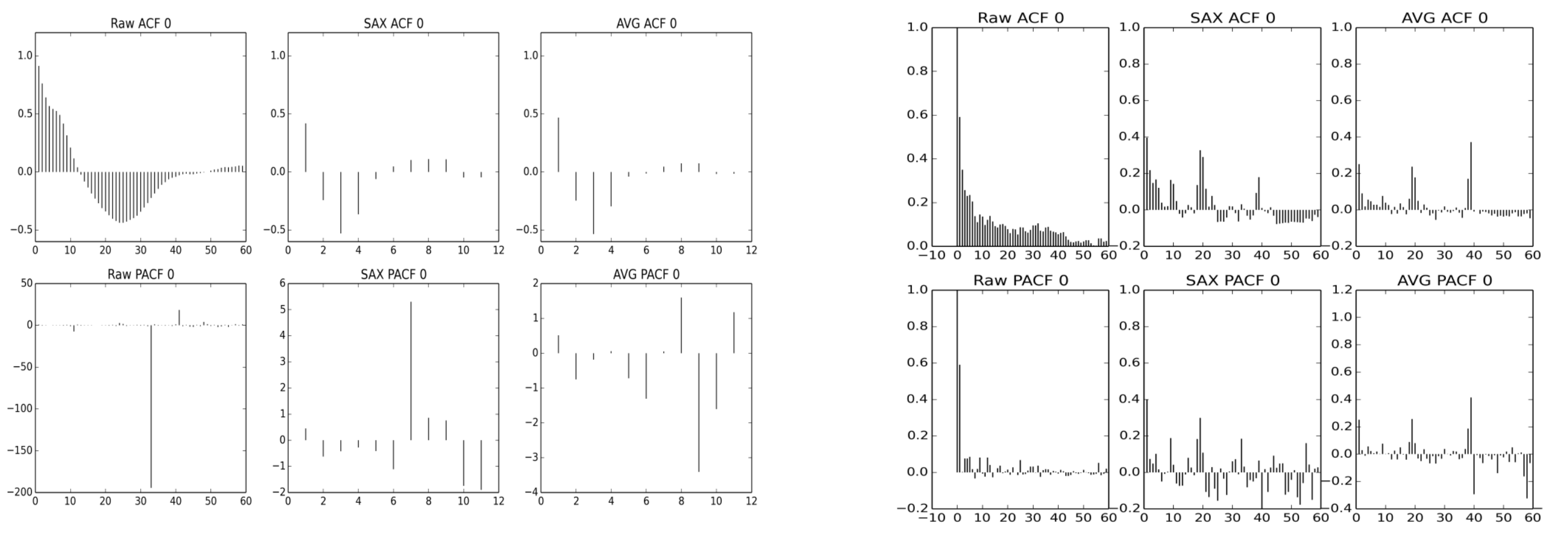}
\caption{ (a) The ACF/PACF plots of raw data, SAX words and PAA output on the benchmark 'ECG' dataset. SAX words and PAA output both preserve the correlation while compressing the original signals. The absolute mean of  ACF value over all samples is 0.35 on the raw signals, 0.23 on the SAX words and 0.21 on PAA values (left), (b) The ACF/PACF plots on the clinical dataset (right).}
\label{fig:ACF-UCR}
\end{figure}

The internal correlation indicates if the representation can preserve the significant seasonal patterns with strong correlation in the data while removing much of the redundant correlations. A good representation should smooth the noise, reduce the volume and simultaneously extract the main patterns. We apply ACF and PACF to evaluate the ability to capture intrinsic correlation embedded in time series on SAX words and PAA output. In our experiments, both PAA and SAX can extract the correlation tendency in the original signals, because the ACF/PACF plots have the same shape on both PAA output and SAX words (Figure \ref{fig:ACF-UCR}). The maximum peak value in ACF measures the periodicity in a sequence of signals \cite{sant2011symbolization}. Thus, we look into the absolute mean among the ACF  values and take the average over all samples in each dataset (Table \ref{tab:ACF-mean}). SAX words actually always have the higher absolute mean ACF than PAA output, thus preserve more internal correlations than PAA output.

We also explored the morphology of the internal correlation embedded in time series by the SAX words and PAA output on the real-world clinical dataset (Figure \ref{fig:ACF-UCR}b). Both SAX words and PAA output have the similar scheme of ACF to preserve the internal correlation. The PACF behaves quite interestingly. The average ACF value of the original signals, SAX words and PAA output are 0.587, 0.556 and 0.360. SAX and PAA both decrease the correlation redundancy to generate more compact representations. The smoothness of the redundancy effect by SAX words also interpret the benefit that SAX compresses the internal correlation.

While the ACF indicates both SAX and PAA will extract the internal correlation, the PACF demonstrates these two methods do not follow the same routines, but have different recurrent relations. In time series analysis, we decide the order of auto-regressive (AR) process from the significance beyond the time lags, because those insignificant coefficients do not exhibit any patterns. Although it is clear that test time series are not a stationary process and most of them have the seasonal trend, the PACF shows complex behavior on SAX words and PAA values. Interestingly, the peak value of PACF on SAX words and PAA output sometimes has the different sign, which means the correlation orientations are opposite after the Gaussian mapping procedure.
\begin{table}[t]
  \centering
  \caption{The absolute mean value of the ACF on raw data, SAX words and PAA output.}
    \begin{tabular}{crrr}
    \toprule
    \textbf{Dataset} & \multicolumn{1}{c}{\textbf{RAW }} & \multicolumn{1}{c}{\textbf{SAX}} & \multicolumn{1}{c}{\textbf{PAA }} \\
    \midrule
    lighting2 & 0.3502 & 0.2336 & 0.2112 \\
    ECG200 & 0.1981 & 0.2907 & 0.2787 \\
    Coffee & 0.4723 & 0.3237 & 0.3227 \\
    Adiac & 0.4656 & 0.3237 & 0.3227 \\
    lighting7 & 0.2455 & 0.2868 & 0.2704 \\
    Beef & 0.4040 & 0.2722 & 0.2599 \\
    Oliveoil & 0.3226 & 0.1763 & 0.1680 \\
    Clinical ECG & 0.6331 & 0.5563 & 0.3605 \\
    \bottomrule
    \end{tabular}%
  \label{tab:ACF-mean}%
\end{table}%

\section{Conclusion}
To the best of our knowledge, this paper is the first attempt to explore the intrinsic statistical properties of Symbolic Aggregation Approximation. We empirically investigates the intrinsic properties of Symbolic Aggregation approXimation method in the statistical perspective. We explain the statistical behavior of SAX words via three aspects, the complexity, the information embedding efficiency and the internal correlation. We also proposed a new measurement, Information Embedding Cost (IEC) to evaluate the information embedding efficiency of the symbolic dynamics. We proved that high correlation lies between the IEC score and the classification performance. 

The logic map is clear. Lower absolute permutation entropy implies that SAX words significantly reduce the complexity of the original time series. This observation motivates us to investigate how much information will the symbolic representations retain and discard after processing by the SAX. We develop a novel measurement, IEC score based on KL divergence and reconstruction information loss to evaluate the information embedding efficiency. Our experiments indicate that SAX shows significant improvement than PAA with lower IEC score.  We recommend the criterion of IEC score to be 0.2 when considering if the SAX representations work better or not. If test representation has lower IEC score than 0.2, it is much likely that SAX will improve the classification performance compared with original signals. Such analysis on IEC score can also be generalized to other symbolic representations such as ACA, Persist or some  newly proposed representation methods.

Redundancy effect incurs in the analysis on permutation entropy. This observation motivate us to investigate how SAX behaves on the internal correlation embedded in the original time series. Our experiments demonstrate that the internal correlation which is measured by ACF is preserved by both SAX and PAA precisely and concisely. However, moving average (MA) model will not fit the output signal well, because the ACF indicates seasonal properties before differencing. The same trends preserved by the ACF and lower average ACF values imply that PAA-based discretization reduce the redundant internal correlation and maintain major correlations. SAX words always have slightly higher average ACF value than PAA, thus maintain the better internal correlations. The PACF behavior are quite different within SAX and PAA, which implies that Gaussian mapping procedure greatly changes the conditional correlation among original data. This observation is interesting. We will try to reveal the symbolic mechanics of this observation in the future work. 

Our work can support the SAX method and its further applications, provide the analytical framework and statistical tools to evaluate symbolic representations methods, and inspire the researcher to design new symbolic dynamics for time series data mining tasks.





\bibliographystyle{IEEEtran}
\bibliography{Bib}
%

\end{document}